%% file: emnlp2020.tex
\pdfoutput=1
\documentclass[11pt,a4paper]{article}
\usepackage[hyphens]{url}
\usepackage[hyperref]{eacl2021}
\usepackage[hyphenbreaks]{breakurl}
\usepackage{times}
\usepackage{latexsym}
\usepackage{float}

\usepackage{textcomp}
\usepackage{physics}
\usepackage{adjustbox}
\usepackage{amsmath}
\usepackage{tikz}
\usepackage{mathdots}
\usepackage{yhmath}
\usepackage{cancel}
\usepackage{xcolor}
\usepackage{siunitx}
\usepackage{array}
\usepackage{multirow}
\usepackage{amssymb}
\usepackage{gensymb}
\usepackage{tabularx}
\usepackage{booktabs}
\usetikzlibrary{fadings}
\usetikzlibrary{patterns}
\usetikzlibrary{shadows.blur}
\usetikzlibrary{shapes}

\usepackage{makecell}
\usepackage{microtype}
\usepackage{tikz}
\usetikzlibrary{shapes,arrows,calc, positioning}
\tikzstyle{block} = [text width=15em, text centered]
\tikzstyle{block2} = [rectangle, draw, fill=gray!10, text width=10em, text centered, rounded corners, minimum height=2em]
\tikzstyle{block3} = [rectangle, draw, fill=gray!0, text width=10em, text centered, rounded corners, minimum height=2em]
\tikzstyle{block4} = [rectangle, draw, fill=gray!0, text width=13em, text centered, rounded corners, minimum height=2em]
\tikzstyle{line} = [draw, -latex']
\tikzstyle{doc}=[%
draw,
thick,
align=center,
color=black,
shape=document,
minimum width=1cm,
minimum height=1cm,
shape=document,
inner sep=2ex,
]
\aclfinalcopy 

\input{figure}

\title{El Volumen Louder Por Favor: Code-switching in Task-oriented Semantic Parsing }

 \author{Arash Einolghozati \quad Abhinav Arora \quad Lorena Sainz-Maza Lecanda \\
\textbf{Anuj Kumar \quad Sonal Gupta}\\
Facebook \\
\texttt{\{arashe,abhinavarora,lorenasml,anujk,sonalgupta\}@fb.com} \\
}

\date{}

\begin{document}
\maketitle
\begin{abstract}
Being able to parse code-switched (CS) utterances, such as Spanish+English or Hindi+English, is essential to democratize task-oriented semantic parsing systems for certain locales. In this work, we focus on Spanglish (Spanish+English) and release a dataset, CSTOP, containing 5800 CS utterances alongside their semantic parses. We examine the CS generalizability of various Cross-lingual (XL) models and exhibit the advantage of  pre-trained XL language models   when data for only one language is present.  As such, we focus on improving the pre-trained models for the case when only English corpus alongside either zero or a few CS training instances are available. We propose two data augmentation methods for the zero-shot and the few-shot settings: fine-tune using translate-and-align and augment using a generation model followed by match-and-filter. Combining the few-shot setting with the above improvements decreases the initial $30$-point accuracy gap between the zero-shot and the full-data settings by two thirds.
\end{abstract}

\section{Introduction}

Code-switching (CS) is the alternation of languages within an utterance or a conversation~\cite{cs_book}.
It occurs under certain linguistic constraints but can vary from one locale to another~\cite{joshi-1982-processing}.  
%
We envision two usages of CS for virtual assistants. First, CS is very common in locales where there is a heavy influence of a foreign language (usually English) in the native “substrate” language (e.g., Hindi or Latin-American Spanish). Second, for other native languages, the prevalence of English-related tech words (e.g., Internet, screen) or media vocabulary (e.g., movie names) is very common. While in the second case, a model using  contextual understanding should be able to parse the utterance, the first form of CS, which is our focus in this paper, needs Cross-Lingual(XL) capabilities in order to infer the meaning.

There are various challenges for CS semantic parsing. First, collecting CS data is hard because it needs bilingual annotators. This gets even worse considering that the number of CS pairs grows quadratically. Moreover, CS is very dynamic and changes significantly by occasion and in time~\cite{cs_book}.
As such, we need extensible solutions that need little or no CS data while having the more commonly-accessible English data available. In this paper, we first focus on the zero-shot setup for which we only use EN data for the same task domains (we call this in-domain EN data). We show that by translating the utterances to ES and aligning the slot values, we can achieve high accuracy on the CS data. Moreover, we show that having a limited number of CS data alongside augmentation with synthetically generated data can significantly improve the performance.

Our contributions are as follows: 1) We release a code-switched task-oriented dialog data set, CSTOP\footnote{The dataset can be downloaded from \href{https://fb.me/cstop_data}{https://fb.me/cstop\_data}}, containing 5800 Spanglish utterances and a corresponding parsing task. To the best of our knowledge, this is the first Code-switched parsing dataset of such size that contains utterances for both training and testing. 2) We evaluate strong baselines under various resource constraints. 3) We introduce two data augmentation techniques that improve the code-switching performance using monolingual data.

\begin{figure}
    \centering
    \small
    \begin{tikzpicture}
    \node [block] (intent) {IN:GET\_WEATHER};
    \node [block, below of=intent, node distance=2.6em] (sentence) {Dime el clima para next Friday};
    \draw[draw=black]  ($(sentence.west)+(0,0.3)$) -- ($(sentence.east)+(0,0.3)$);
    \draw[draw=black]  ($(sentence.west)-(-2.35,0.3)$) -- ($(sentence.east)-(0.4,0.3)$);
    \node [block, below of=sentence, node distance=2.5em, xshift=3em] (slot2) {SL:DATE\_TIME};
    \path [line] (sentence.north) -| (intent.south);
    \path [line] ($(sentence.south)+(.95,0)$) -| (slot2.north);
    \end{tikzpicture}
    \caption{Example CS sentence and its annotation for the sequence [IN:GET\_WEATHER Dime el clima [SL:DATE\_TIME para next Friday]]}
    \label{fig:example}
\end{figure}
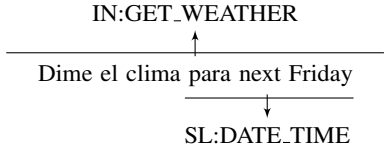

\section{Task}
In task-oriented dialog, the language understanding task consists of classifying the intent of an utterance, i.e., sentence classification, alongside tagging the slots, i.e., sequence labeling. 
We use the Task-Oriented Parsing dataset released by~\citet{sebastianMultilingual} as our EN monolingual dataset. 
We release a similar dataset, CSTOP, of around 5800 Spanglish utterances over two domains, Weather and Device, which are collected and annotated by native Spanglish speakers. 
An example from the CSTOP alongside its annotation is shown in Fig.~\ref{fig:example}. Note that the intent and slot lables start with $IN:$ and $SL:$, respectively. Our task is to classify the sentence intent, here \textit{IN:GET\_WEATHER} as well as the label and value of the slots, here \textit{SL:DATE\_TIME} corresponding to the span \textit{para next Friday}. Moreover, other words are classified as having no label, i.e., \textit{O} class. We discuss the details of this dataset in the next section.

One of the unique challenges of this task, compared with common NER and language identification CS tasks, is the constant evolution of CS data. Since the task is concerned with spoken language, the nature of CS is very dynamic and keeps evolving from domain to domain and from one community to another. Furthermore, cross-lingual data for this task is also very rare. Most of the existing techniques, either combine monolingual representations~\cite{winata-etal-2019-hierarchical} or combine the datasets to synthesize code-switched data~\cite{liu2019attentioninformed}. Lack of monolingual data for the substrate language (very realistic if you replace ES with a less common language) would make those techniques inapplicable.

In order to evaluate the model in a task-oriented dialog setting, we use the exact-match accuracy (from now on, accuracy) as the primary metric. This is simply defined as the percentage of utterances for which the full parse, i.e., the intent and all the slots, have been correctly predicted. 

\section{CSTOP Dataset}
In this section, we provide details of the CSTOP dataset. We originally collected around 5800 CS utterances over two domains; \textit{Weather} and \textit{Device}. We picked these two domains as they represent complementary behavior. While Weather contains slot-heavy utterances (average 1.6 slots per utterance), Device is an intent-heavy domain with only average 0.8 slots per utterance. We split the data into 4077, 1167, and 559 utterances for training, testing, and validation, respectively.

CS data collection proceeded in the following steps:
\begin{enumerate}
    \item One of the authors, who is a native speaker of Spanish and uses Spanglish on a daily basis, generated a small set of CS utterances for Weather and Device domains. Additionally, we also recruited bilingual EN/ES speakers who met our Spanglish speaker criteria guidelines, established following ~\citet{spanglish_book}.
    \item We wrote Spanglish data creation instructions and asked participants to produce Spanish-English CS utterances for each intent (i.e. ask for the weather, set device brightness, etc).
    \item Next, we filter out utterances from this pool to only retain those that exhibited true intra-sentential CS.
    \item The collected utterances were labeled by two annotators, who identified the intent and slot spans. If the two annotators disagreed on the annotation for an utterance, a third annotator would resolve the disagreement to provide a final annotation for it.
\end{enumerate}


Table.~\ref{tab:data_stats} shows the number of distinct intents and slots for each domain and the number of utterances in CSTOP for each domain. We have also shown the most 15 common intents in the training set and a representative Spanglish example alongside its slot values for those intents in Table.~\ref{tab:data_examples}. The first value in a slot tuple is the slot label and the second is the slot value.
We can see that while most of the verbs and stop words are in Spanish, Nouns and slot values are mostly in English. We further calculate the prevalence Spanish and English words by using a vocabulary file of 20k for each language. Each token in the CSTOP training set is assigned to the language for which that token has a lower rank. The ratio of the Spanglish to English tokens is around $1.34$ which matches our previous anecdotal observation. This ratio was consistent when increasing the vocabulary size to even 40k. 
.
\begin{table}
\centering
\begin{tabular}{lccc}
\hline
\textbf{Domain} & \textbf{\# intents} & \textbf{\# slots} & \textbf{\#  utterances}\\
Weather &  2 & 4 & 3692\\
Device & 17 & 6 & 2112\\
\hline
\end{tabular}
\caption{CSTOP Statistics}
\label{tab:data_stats}
\end{table}

\begin{table*}
\small
\centering
\begin{tabular}{lcc}
\hline
\textbf{intent} &  \textbf{utterance} & \textbf{slots}\\
GET\_WEATHER & ¿cómo estará el clima en Miami este weekend?  & \makecell{(LOCATION, Miami), \\ (DATE\_TIME, este weekend)}\\
\hline
UNSUPPORTED\_WEATHER &  how many centimeters va a llover hoy & (DATE\_TIME, hoy)\\
\hline
OPEN\_RESOURCE &  Abreme el gallery & (RESOURCE, el gallery)\\
\hline
CLOSE\_RESOURCE  & Cierra maps & (RESOURCE, maps)\\
\hline
TURN\_ON & Prende el privacy mode & (COMPONENT, el privacy mode) \\
\hline
TURN\_OFF & Desactiva el speaker & (COMPONENT, el speaker)\\
\hline
WAKE\_UP  &  Quita sleep mode & -\\
\hline
SLEEP  & prende el modo sleep & -\\
\hline
OPEN\_HOMESCREEN  &  Go to pagina de inicio & -\\
\hline
MUTE\_VOLUME &    Desactiva el sound & -\\
\hline
UNMUTE\_VOLUME  &  Prende el sound & -\\
\hline
SET\_BRIGHTNESS &  subir el brigtness al 80 & (PERCENT, 80) \\
\hline
INCREASE\_BRIGHTNESS  &  Ponlo mas bright & -\\
\hline
DECREASE\_BRIGHTNESS  & baja el brightness  & -\\
\hline
SET\_VOLUME &  Turn the volumen al nivel 10 & (PRECISE\_AMOUNT,10)\\
\hline
INCREASE\_VOLUME  &  aumenta el volumen a little bit & -\\
\hline
DECREASE\_VOLUME &   Bájale a la music & -\\
\hline
\end{tabular}
\caption{Examples from CSTOP intents }
\label{tab:data_examples}
\end{table*}

\section{Model}
Our base model is a bidirectional LSTM with separate projections for the intent and slot tagging~\cite{slotFillingYao}. We use the aligned word embedding MUSE~\cite{muse} with a vocabulary size of 25k for both EN and ES.  Our experiments showed that for the best XL generalization, it’s best to freeze the word embeddings when the training data contains only EN or ES utterances. We refer to this model as simply MUSE.

We also use SOTA pre-trained XL models; XLM~\cite{XLM} and XLM-R~\cite{XLM_r}. These models are pre-trained via Masked Language Modeling (MLM)~\cite{bert} on massive multilingual data. They share the word-piece token representation, BPE~\cite{bpe_sennrich} and SentencePiece~\cite{sentencepiece}, as well as a common MLM transformer for different languages. 
Moreover, while XLM is pre-trained on Wikipedia, XLM-R is trained on crawled web data which contains more non-English and possibly CS data. 
In order to adapt these models for the joint intent classification and slot tagging task, we use the method described in~\citet{bert_joint}. For classification, we add a linear classifier on top of the first hidden state of the Transformer. A typical slot tagging model feeds the hidden states, corresponding to each token, to a CRF layer~\cite{slotFillingMesnil}. To make this compatible with XLM and XLM-R, we use the hidden states corresponding to the first sub-word of every token as the input to the CRF layer. 

Table~\ref{tab:zero} shows the accuracy of the above models on CSTOP. 
 We also have listed the performance when the models were first fine-tuned on the EN data (CS+EN). We observe that in-domain fine-tuning can almost halve the gap between XLM-R and XLM, which is around $50\%$ faster during the inference than XLM-R during inference. The training details for all our models and the validation results are listed in the Appendix.

\section{Zero-shot performance}

Bottom part of Table~\ref{tab:zero} shows the CS test accuracy when using only the in-domain monolingual data. Our EN dataset is the task-oriented parsing dataset~\cite{sebastianMultilingual} described in the previous section. Since the original TOP dataset did not include any utterances belonging to the Device domain, we also release a dataset of around thousand EN Device utterances for the experiments using the EN data. In order to showcase the effect of monolingual ES data, we also experiment with using the in-domain ES dataset, i.e. ES Weather and Device queries.

\begin{table}
\centering
\begin{tabular}{lccc}
\hline
\textbf{ Lang/Model} &\textbf{MUSE} & \textbf{XLM} & \textbf{XLM-R} \\
CS & 87.0  & 86.6 & 94.4 \\ 
CS + EN & 88.1& 93.0 & 95.4 \\
\hline
EN & 39.2 & 54.8 &  66.6 \\
ES & 69.9 &  78.3 & 88.1\\
EN+ES & 88.2 & 87.8 & 91.2\\
\hline
\end{tabular}
\caption{Full-training (top) and zero-shot (bottom) accuracy of XL models when using different monolingual corpora. ES is an internal dataset to showcase the effect of having a big Spanish corpus.}
\label{tab:zero}
\end{table}

We observe that having monolingual data of both languages yields very high accuracy, only a few points shy of training directly on the CS data. 
Moreover, in this setting, even simpler models such as MUSE can yield competitive results with XLM-R while being much faster. 
However, the advantage of XL pre-training becomes evident when only one of the languages is present. As such, having only the substrate language (i.e., ES) is almost the same as having both languages for XLM-R.  

Note that we do not use ES data for other results in this paper. Obtaining semantic parsing dataset in another language is expensive and often only EN data is available. Our experiments show a huge performance gap when only using the EN data, and thus in this paper, we will be focusing on using the EN data alongside zero or a few CS instances.


\subsection{Effect of XL Embeddings} 
Here, we explore how much of the zero-shot performance can be attributed to the XL embeddings as opposed to the shared XL representation. 
As such, we experiment with replacing MUSE embeddings with other embeddings in the LSTM model explained in the previous section.  
We experiment with the following strategies::
(1) Random embedding: This learns the ES and EN word embeddings from the scratch
(2) Randomly-initialized SentencePiece~\cite{sentencepiece} (RSP): Words are represented by wordpiece tokens that are learned from a huge unlabeled multilingual corpus. 
(3) Pre-trained XLM-R sentence piece (XLSP). These are the 250k embedding vectors that are learned during the pre-trainig of XLM-R. 

We have shown the effects of using the aforementioned embeddings in the zero-shot setting in Table~\ref{tab:zero_embedding}. 
We can see that by having monolingual datasets from both languages, even random embeddings can yield high performance.
By removing one of the languages, unsurprisingly, the codeswitching generalizability  drops sharply for all, but much less for XLSP and MUSE.  Moreover, even though the XLSP embeddingsm, unlike MUSE, is not consttrained to only EN and ES, it yields comparable results with the word-based MUSE embeddings.

We can also see that When ES data is available, RSP provides some codeswitching generalizability, as compared with the Random strategy, but not when only EN data is available. We hypothesize that the common sub-word tokens are more helpful to generalize the slot values (which in the codeswitched data are mostly in EN) than the non-slot queries which are more commonly in ES. This is also verified by the observation that most of the gains for the RSP vs Random for the ES only scenario come from the slot tagging accuracy as compared with the intent detection.

As a final note, we observe that between $20-30\%$ of the XLM-R gains can be captured by using the pre-trained sentence-piece embeddings while the rest are coming from the shared XL representation pre-trained on massive unlabeled data.
In the rest of the paper, we focus on the XLM-R model.

\begin{table}
\centering
\begin{tabular}{lcccc}
\hline

 &\textbf{Random} & \textbf{RSP} & \textbf{XLSP} & \textbf{MUSE} \\
EN & 13.5 & 12.2 &  30.3 &  \bf{39.2}\\
ES & 38.2 & 48.0 & \bf{70.5} & \bf{69.9}\\
EN+ES & 81.1 & 84.3 & \bf{89.0} & 88.2\\
\hline
\end{tabular}
\caption{Zero-shot accuracy for simple LSTM model when using different monolingual corpora and different embedding strategies.}
\label{tab:zero_embedding}
\end{table}

\insertTranslateExample

\section{Data Augmentation Approaches}
In this section, we discuss two data augmentation approaches. The first one is in a zero-shot setting and only uses EN data to improve the performance on the Spanglish test set. In the second approach, we assume having a limited number of Spanglish data and use the EN data to augment the few-shot setting. 

\subsection{Translate and Align}
We explore creating synthetic ES data from the EN dataset using machine translation. Since our task is a joint intent and slot tagging task, creating a synthetic ES corpus consists of two parts: a) Obtaining a parallel EN-ES corpus by machine translating utterances from EN to ES, b) Projecting gold annotations from EN utterances to their ES counterparts via word alignment~\cite{tiedemann-etal-2014-treebank,lee-etal-2019-learning}. Once the words in both languages are aligned, the slot annotations are simply copied over from EN to ES by word alignment. For word alignment, we explore two methods that are explained below. In some cases, word alignment may produce discontinuous slot tokens in ES, which we handle by introducing new slots of the same type, for all discontinuous slot fragments.

Our first method leverages the attention scores~\cite{BahdanauCB14} obtained from an existing EN to ES NMT model. We adopt a simplifying assumption that each source word is aligned to one target language word \cite{brown-etal-1993-mathematics}. For every slot token in the source language, we select the highest attention score to align it with a word in the target language.

Our next approach to annotation projection makes use of unsupervised word alignment from statistical machine translation. Specifically, we use the fast-align toolkit~\cite{dyer-etal-2013-simple} to obtain alignments between EN and ES tokens. Since fast-align generates asymmetric alignments, we generate two 
sets of alignments, EN to ES and ES to EN and symmetrize them using the grow-diagnol-final-and heuristic~\cite{koehn-etal-2003-statistical} to obtain the final alignments.

\begin{figure*}[t!]
\centering
\small
\makebox[0pt]{
    \begin{tikzpicture}[node distance = 1cm, scale=1, every node/.style={scale=1}, auto]
        
        \node [block2] (esen) {[IN:GET\_WEATHER show me the weather [SL:DATE\_TIME  for next Monday ]};
        
        \node [block3, left of=esen, node distance=5cm]  (orig){ [IN:GET\_WEATHER Dime el clima  [SL:DATE\_TIME para next Friday]};
        
        \node [block4, right of=esen, node distance=5cm, yshift=1cm] (beam1) {[IN:GET\_WEATHER Quiero saber el clima  [SL:DATE\_TIME  para next Monday ]]};
        \node [block4, right of=esen, node distance=5cm, yshift=-.8cm] (beam2) {[IN:GET\_WEATHER Dime el clima esperado [SL:DATE\_TIME  para next Friday ]]};
        \node [block4, right of=esen, node distance=5cm, yshift=-2.5cm] (beam3) {[IN:GET\_WEATHER Dime el pronóstico [SL:DATE\_TIME hasta el 15 ]]};
        Draw edges
       
        \path [line] (esen) -- (beam1.west);
        \path [line] (esen) -- (beam2.west);
        \path [line] (esen) -- (beam3.west);
        \path [line] (orig) -- (esen.west);
    \end{tikzpicture}
}
\caption{Match and Filter data augmentation: 1- For each CS utterance (target), find the the closest EN neighbor (source). 2- Learn a generative model from source to target 3- Perform beam search to generate more targets from the source utterances.}
\label{fig:augment}
\end{figure*}
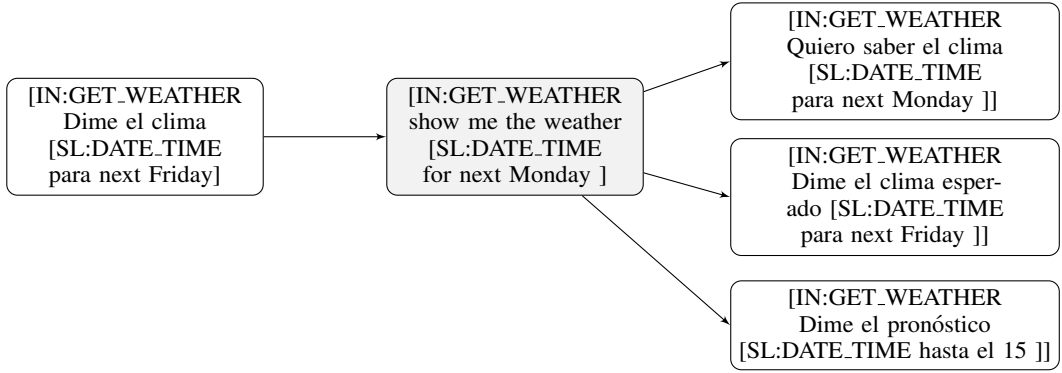

In Table~\ref{tab:zero_translate}, we show the CS zero-shot accuracy when fine-tuning on the newly generated ES data (called $\textrm{ES}^*$.) alongside the original EN data. We can see that unsupervised alignment results in around 2.5 absolute point accuracy improvement. On the other hand, using attention alignment ends up hurting the accuracy, which is perhaps due to the slot noise that it introduces. The assumption that a single source token aligns with a single target token leads to incorrect data annotations when the length of a translated slot is different in EN and ES.  Figure~\ref{fig:augment} shows an example utterance where attention alignment produces an incorrect annotation compared to unsupervised alignment.

\begin{table}[H]
\centering

\begin{tabular}{lcc}
\hline
\textbf{ EN} & \bf{EN+$\textrm{ES}^*$ Attn} & \bf{EN+$\textrm{ES}^*$ aligned} \\
 66.6     & 65.8 & 69.2 \\
\hline
\end{tabular}
\caption{Zero-shot accuracy when fine-tuning XLM-R on EN monolignual data as well as the  auto-translated and aligned ES data (called ES*).}
\label{tab:zero_translate}
\end{table}

\begin{table*}
\centering
\begin{tabular}{lcc}
\hline
\textbf{Model/Training Data} & \textbf{Few Shot} & \textbf{Few shot+ Generate and Filter augmentation} \\
XLM-R &  61.2 & 70.3\\
XLM-R fine-tuned on EN & 82.6   & 83.7\\
XLM-R fine-tuned on EN+$\textrm{ES}^*$  &84.1 &  \bf{84.8}\\
\hline
\end{tabular}
\caption{Accuracy when only a few CS instances are available during training, with and without the data augmentation. ES* is the auto-translated and aligned data.}
\label{tab:few}
\end{table*}

\subsection{Generate by Match-and-Filter in the Few-shot Setting}

Here, we assume having a limited number of high-quality in-domain CS data and as such, we construct the $\textrm{CSTOP}_\textrm{100}$ dataset of around $100$ utterances from the original training set in the CSTOP. We make sure that every individual slot and intent (but not necessarily the combination) is presented in $\textrm{CSTOP}_\textrm{100}$  and randomly sample the rest. We perform our sampling three times and report the few-shot results on the average performance.  This setting is of paramount importance for bringing up a domain in a new locale when the EN data is already available.
The first column in Table~\ref{tab:few} shows the CS Few-Shot (FS) performance alongside the fine-tuning on the EN data and the aligned translated data, when average over three sampling of $\textrm{CSTOP}_\textrm{100}$.

In order to improve the FS performance,  we perform data augmentation on the $\textrm{CSTOP}_\textrm{100}$ dataset. Unlike methods such as ~\citet{cs_rule_data}, we seek generic methods that do not need extra resources such as constituency parsers. Instead, we explore using pre-trained generative models while taking advantage of the EN data.

We use BART~\cite{bart}, a denoising autoencoder trained on massive amount of web data, as the generative model. Our goal is to generate diverse Spanglish data from the EN data. Even though BART was trained for English, we found it very effective for this task. We hypothesize this is due to the abundance of the Spanish text among EN web data and the proximity of the word-piece tokens among them. We also experimented with multilingual BART~\cite{mbart} but found it very challenging to fine-tune for this task.

First, we convert the data to a bracket format~\cite{grammar}, which is called the seqlogical form in~\citet{gupta2018rnng}. Examples of this format are shown in Fig.~\ref{fig:augment}. In the seqlogical form, we include the intent (i.e., sentence label) at the beginning and for each slot, we first include the label and text in brackets.

We perform our data augmentation technique in the following steps: 
\begin{enumerate}
    \item  Find the top K closest EN neighbors to every CS query in the  $\textrm{CSTOP}_\textrm{100}$. We enforce the neighbors to have the same parse as the CS utterance, i.e., same intent and same slot labels, and use the Levenshtein distance to rank the EN sequences. 
    
\item Having this parallel corpus, i.e., top-K EN neighbors as the source and the original CS query as the target, Fine-tune the BART model. We use K=10 in our experiments to increase the parallel data size to around $650$.

\item During the inference, Use the beam size of 5 to decode CS utterances from the same EN source data. Since both the source and target sequences are in the seqlogical form, the CS generated sequences are already annotated.

\end{enumerate}

In Fig.~\ref{fig:augment}, we have shown the closest EN neighbor corresponding to the original CS example in Fig.~\ref{fig:example}. The  CS utterance can be seen as a rough translation of the EN sentence. We have also shown the top three generated CS utterances from the EN example.

In order to reduces the noise, we filter the generated sequences that either already exist in $\textrm{CSTOP}_\textrm{100}$, are not valid trees, or  have a semantic parse different from the original utterance.  We augment $\textrm{CSTOP}_\textrm{100}$  with the data, and fine-tune the XLM-R baseline.

In the second column of Table~\ref{tab:few}, we have shown the average data augmentation improvement over the three $\textrm{CSTOP}_\textrm{100}$ samples for the few-shot setting. We can see that even after fine-tuning on the EN monolingual data (the second row), the augmentation technique improves this strong baseline. In the last row, we first use the translation alignment of the previous section to obtain  $\textrm{ES}^*$. After fine-tuning on this set combined with the EN data, we further fine-tune on the $\textrm{CSTOP}_\textrm{100}$.  We can see that the best model enjoys improvements from both zero-shot (translation alignment) and the few-shot (generate and filter) augmentation techniques. We also note that the p-value corresponding to the second and third row gains are 0.018 and 0.055, respectively.

\section{Related Work}
\subsection{XL Pre-training}
Most of the initial work on pre-trained XL representations was focused on embedding alignment~\cite{xing-etal-2015-normalized,zhang-etal-2017-adversarial,muse}. Recent developments in this area have focused on the context-aware XL alignment of contextual representations~\cite{Schuster2019cross, aldarmaki-diab-2019-context, wang-etal-2019-cross,iclr/CaoKK20}. Recently, pre-trained multilingual language models such as mBERT~\cite{bert}, XLM~\cite{XLM}, and~\citet{XLM_r}  have been introduced, and~\citet{PiresSG19} demonstrate the effectiveness of these on sequence labeling tasks.\\
Separately,~\citet{mbart} introduce mBART, a sequence-to-sequence denoising auto-encoder pre-trained on monolingual corpora in many languages using a denoising autoencoder objective \citep{bart}.

\subsection{Code-Switching}
Following the ACL shared tasks, CS is mostly discussed in the context of word-level language identification~\cite{cs_lid} and NER~\cite{cs_ner}. Techniques such as curriculum learning~\cite{cs_curriculum} and attention over different embeddings~\cite{cs_ner_attention, winata-etal-2019-hierarchical} have been among the successful techniques.
CS parsing and use of monolingual parses are discussed in~\citet{cs_shallow,cs_join,cs_universal}.
\citet{cs_shallow} introduces a Hinglish test set for a shallow parsing pipeline. In~\citet{cs_join}, outputs of two monolingual dependency parsers are combined to achieve a CS parse.~\citet{cs_universal} extends this test set by including training data and transfers the knowledge from monolingual treebanks.
\citet{multi_cs} introduced a CS test set  for semantic parsing which is curated by combining  utterances from the two monolingual datasets.  In contrast, CSTOP is procured independently of the monolingual data and exhibits much more linguistic  diversity.
In~\citet{cs_rule_data}, linguistic rules are used to generate CS data which has been shown to be effective in reducing the perplexity  of a CS language model. In contrast, our augmentation techniques are generic and do not require rules or constituency parsers.\\
\subsection{XL Data Augmentation}
Most approaches to cross-lingual data augmentation use machine translation and slot projection for sequence labeling tasks~\citep{jain-etal-2019-entity}.
\citet{eda} uses simple operations such as synonym replacement and \citet{LeeY019} use phrase replacement from a parallel corpus to augment the training data.
~\citet{xlda} present XLDA that augments data by replacing segments of input text with its translations in other languages.
Some recent approaches \citep{ChangCL19,WinataMWF19} also train generative models to artificially generate CS data. More recently,~\citet{abs-2003-02245} study data augmentation using pre-trained transformer models by incorporating label information during fine-tuning. Concurrent to our work,~\citet{multimix} introduce Multimix, where data augmentation from pre-trained multilingual language models and self-learning are used for semi-supervised learning. Recently,~\citet{liu2019attentioninformed} generate CS data by translating keywords picked based on attention scores from a monolingual model.
Generating CS data has recently been studied in~\citet{Liu_Winata_Lin_Xu_Fung_2020}

\subsection{Task-oriented Dialog}
The intent/slot framework is the most common way of performing language understanding for task oriented dialog using. Bidirectional LSTM for the sentence representation alongside separate projection layers for intent and slot tagging is the typical architecture for the joint task~\cite{slotFillingYao,slotFillingMesnil,HakkaniTur2016}. Such representations can accommodate trees of up to length two, as is the case in CSTOP. More recently, an extension of this framework has been introduced to fit the deeper trees~\cite{gupta2018rnng, alexa_seq2seq}. 

\section{Conclusion}
In this paper, we propose a new task for code-switched semantic parsing and release a dataset, CSTOP, containing 5800 Spanglish utterances over two domains. We hope this foments further research on the code-switching phenomenon which has been set back by paucity of sizeable curated datasets.
We show that cross-lingual pre-trained models can generalize better than traditional models to the code-switched setting when monolingual data from only one languages is available.  In the presence of only EN data, we introduce generic augmentation techniques based on translation and generation. As such, we show that translating and aligning the EN data can significantly improve the zero-shot performance. Moreover, generating code-switched data using a generation model and a match-and-filter approach leads to improvements in a few-shot setting. We leave exploring and combining other augmentation techniques to future work.

\bibliography{emnlp2020}
\bibliographystyle{acl_natbib}

\clearpage

\appendix

\section{Appendix}
Here, we describe the details regarding the training as the validation results.

\subsection{Model and Training Parameters}
In Table~\ref{tab:training_params}, we have shown the training details for all our models. We use ADAM~\cite{adam} with Learning Rate (LR), Weight Decay (WD), and Batch Size (BSz) that is listed for each model. We have also shown the number of epochs and the average training time for the full CS data using $8$ V100 Nvidia GPUs.
 For all our XLM-R experiments, we use the XLM-R large from  the PyText\footnote{\url{https://pytext.readthedocs.io/en/master/xlm\_r.html}}~\cite{pytext} which is pre-trained on 100 languages.   
For the XLM experiments, we use XLM-20 pre-trained over 20 languages and use the same fine-tuning parameters as XLM-R but run for more epochs.

For the LSTM models, we use a two-layer LSTM with hidden dimension of $256$ and dropout of $0.3$ for all connections. We use one layer of MLP of dimension $200$ for both the slot tagging and the intent classification. 
We also use an ensemble of five models for all the LSTM experiments to reduce the variance. The LSTM model with SentencePiece embeddings in Table~\ref{tab:zero_embedding} were trained with embedding dimension of $1024$ similar to the XLM-R model.

\subsection{Validation Results}                    
In Table.~\ref{tab:full_eval}, we have shown the validation results when using the full CS training data. We have not shown the corresponding results for the zero-shot experiments as no validation data was not used and the monolingual models were tested off the shelf. 

In Table.~\ref{tab:few_eval}, we have shown the validation results for the few-shot setting.

\begin{table*}
\centering
\begin{tabular}{lccccc}
\hline
Model & BSz & LR & WD  & Epoch & Avg Time  \\
XLM-R (pronoun) & 8 &  0.000005 & 0.0001 & 15  & 5 hr \\
XLM (pronoun) & 8 &  0.000005 & 0.0001 & 20  &  1 hr \\
LSTM (pronoun+question) & 64 & 0.03 & 0.00001 & 45 & 45 min \\
\hline
\end{tabular}
\caption{Training Parameters}
\label{tab:training_params}
\end{table*}

\begin{table*}[t]
\centering
\begin{tabular}{lcc}
\hline
\textbf{Model/Training Data} & \textbf{Few shot} & \textbf{Few shot + Generate and Filter Augmentation} \\
XLM-R & 61.7 & 70.4  \\
XLM-R fine-tuned on EN & 83.3 & 83.9 \\
XLM-R fine-tuned on EN+$\textrm{ES}^*$ & 83.5 & 84.9  \\
\hline
\end{tabular}
\caption{Validation Accuracy when only a few CS instances (FS) are available during training. FS+G refers to augmenting the few-shot instances with generated CS data. ES* is the auto-translated and aligned data.}
\label{tab:few_eval}
\end{table*}

\begin{table*}
\centering
\begin{tabular}{lccc}
\hline
\textbf{ Lang/Model} &\textbf{MUSE} & \textbf{XLM} & \textbf{XLM-R} \\
CS & 87.8 & 90.7 & 95.0\\ 
CS + EN & 89.0 & 92.9 & 95.5  \\
\hline
\end{tabular}
\caption{Validation results for the Full-training on the CS data}
\label{tab:full_eval}
\end{table*}

\end{document}

%% file: figure.tex
\newcommand{\insertTranslateExample}{
\begin{figure*}

\centering

\tikzset{every picture/.style={line width=0.75pt}} 

\begin{tikzpicture}[x=0.75pt,y=0.75pt,yscale=-1,xscale=1]

\draw    (103.5,83.75) -- (126.5,138.75) ;
\draw    (31,82.75) -- (29,140.75) ;
\draw    (134,82.75) -- (154,137.75) ;
\draw    (174.5,83.25) -- (221,139.75) ;
\draw    (72.01,81.75) -- (90,141.75) ;
\draw  [fill={rgb, 255:red, 181; green, 208; blue, 240 }  ,fill opacity=1 ] (14.63,60.07) -- (43.92,60.07) -- (43.92,82.16) -- (14.63,82.16) -- cycle ;
\draw  [fill={rgb, 255:red, 181; green, 208; blue, 240 }  ,fill opacity=1 ] (48.52,59.92) -- (90.04,59.92) -- (90.04,82.01) -- (48.52,82.01) -- cycle ;
\draw  [fill={rgb, 255:red, 181; green, 208; blue, 240 }  ,fill opacity=1 ] (94.27,59.92) -- (116.89,59.92) -- (116.89,82.01) -- (94.27,82.01) -- cycle ;
\draw  [fill={rgb, 255:red, 220; green, 226; blue, 234 }  ,fill opacity=1 ] (121.22,59.92) -- (144.58,59.92) -- (144.58,82.01) -- (121.22,82.01) -- cycle ;
\draw  [fill={rgb, 255:red, 220; green, 226; blue, 234 }  ,fill opacity=1 ] (152.08,59.77) -- (191,59.77) -- (191,81.86) -- (152.08,81.86) -- cycle ;
\draw  [fill={rgb, 255:red, 245; green, 203; blue, 208 }  ,fill opacity=1 ] (7.5,138.37) -- (42.48,138.37) -- (42.48,160.46) -- (7.5,160.46) -- cycle ;
\draw  [fill={rgb, 255:red, 245; green, 203; blue, 208 }  ,fill opacity=1 ] (48.08,138.37) -- (67.36,138.37) -- (67.36,160.46) -- (48.08,160.46) -- cycle ;
\draw  [fill={rgb, 255:red, 245; green, 203; blue, 208 }  ,fill opacity=1 ] (71.5,138.52) -- (117.89,138.52) -- (117.89,160.61) -- (71.5,160.61) -- cycle ;
\draw  [fill={rgb, 255:red, 245; green, 203; blue, 208 }  ,fill opacity=1 ] (121.85,138.67) -- (138,138.67) -- (138,159.75) -- (121.85,159.75) -- cycle ;
\draw  [fill={rgb, 255:red, 244; green, 231; blue, 232 }  ,fill opacity=1 ] (143.84,138.67) -- (163.12,138.67) -- (163.12,159.75) -- (143.84,159.75) -- cycle ;
\draw  [fill={rgb, 255:red, 245; green, 203; blue, 208 }  ,fill opacity=1 ] (167.02,138.52) -- (195.2,138.52) -- (195.2,160.61) -- (167.02,160.61) -- cycle ;
\draw  [fill={rgb, 255:red, 244; green, 231; blue, 232 }  ,fill opacity=1 ] (200.17,138.37) -- (236,138.37) -- (236,160.46) -- (200.17,160.46) -- cycle ;
\draw    (382.5,83.75) -- (405.5,138.75) ;
\draw    (310,82.75) -- (308,140.75) ;
\draw    (413,82.75) -- (433,137.75) ;
\draw    (455,82.75) -- (500,139.75) ;
\draw    (351.01,81.75) -- (369,141.75) ;
\draw  [fill={rgb, 255:red, 181; green, 208; blue, 240 }  ,fill opacity=1 ] (293.63,60.07) -- (322.92,60.07) -- (322.92,82.16) -- (293.63,82.16) -- cycle ;
\draw  [fill={rgb, 255:red, 181; green, 208; blue, 240 }  ,fill opacity=1 ] (327.52,59.92) -- (369.04,59.92) -- (369.04,82.01) -- (327.52,82.01) -- cycle ;
\draw  [fill={rgb, 255:red, 181; green, 208; blue, 240 }  ,fill opacity=1 ] (373.27,59.92) -- (395.89,59.92) -- (395.89,82.01) -- (373.27,82.01) -- cycle ;
\draw  [fill={rgb, 255:red, 220; green, 226; blue, 234 }  ,fill opacity=1 ] (400.22,59.92) -- (423.58,59.92) -- (423.58,82.01) -- (400.22,82.01) -- cycle ;
\draw  [fill={rgb, 255:red, 220; green, 226; blue, 234 }  ,fill opacity=1 ] (431.08,59.77) -- (470,59.77) -- (470,81.86) -- (431.08,81.86) -- cycle ;
\draw  [fill={rgb, 255:red, 245; green, 203; blue, 208 }  ,fill opacity=1 ] (286.5,138.37) -- (321.48,138.37) -- (321.48,160.46) -- (286.5,160.46) -- cycle ;
\draw  [fill={rgb, 255:red, 245; green, 203; blue, 208 }  ,fill opacity=1 ] (327.08,138.37) -- (346.36,138.37) -- (346.36,160.46) -- (327.08,160.46) -- cycle ;
\draw  [fill={rgb, 255:red, 245; green, 203; blue, 208 }  ,fill opacity=1 ] (350.5,138.52) -- (396.89,138.52) -- (396.89,160.61) -- (350.5,160.61) -- cycle ;
\draw  [fill={rgb, 255:red, 245; green, 203; blue, 208 }  ,fill opacity=1 ] (400.85,138.67) -- (417,138.67) -- (417,159.75) -- (400.85,159.75) -- cycle ;
\draw  [fill={rgb, 255:red, 244; green, 231; blue, 232 }  ,fill opacity=1 ] (422.84,138.67) -- (442.12,138.67) -- (442.12,159.75) -- (422.84,159.75) -- cycle ;
\draw  [fill={rgb, 255:red, 244; green, 231; blue, 232 }  ,fill opacity=1 ] (446.02,138.52) -- (474.2,138.52) -- (474.2,160.61) -- (446.02,160.61) -- cycle ;
\draw  [fill={rgb, 255:red, 244; green, 231; blue, 232 }  ,fill opacity=1 ] (479.17,138.37) -- (515,138.37) -- (515,160.46) -- (479.17,160.46) -- cycle ;
\draw    (350.5,81.75) -- (337,138.75) ;
\draw    (454.5,81.75) -- (460.5,138.25) ;

\draw (19.9,66.17) node [anchor=north west][inner sep=0.75pt]  [font=\footnotesize] [align=left] {{\small set}};
\draw (49.37,65.17) node [anchor=north west][inner sep=0.75pt]  [font=\footnotesize] [align=left] {{\small volume}};
\draw (99.34,64.17) node [anchor=north west][inner sep=0.75pt]  [font=\footnotesize] [align=left] {{\small to}};
\draw (124.91,65.17) node [anchor=north west][inner sep=0.75pt]  [font=\footnotesize] [align=left] {{\small 10}};
\draw (10.4,142.93) node [anchor=north west][inner sep=0.75pt]  [font=\footnotesize] [align=left] {{\small ajuste}};
\draw (51.82,142.78) node [anchor=north west][inner sep=0.75pt]  [font=\footnotesize] [align=left] {{\small el}};
\draw (73.17,142.93) node [anchor=north west][inner sep=0.75pt]  [font=\footnotesize] [align=left] {{\small volumen}};
\draw (125.67,143.93) node [anchor=north west][inner sep=0.75pt]  [font=\footnotesize] [align=left] {{\small a}};
\draw (146.66,143) node [anchor=north west][inner sep=0.75pt]  [font=\footnotesize] [align=left] {{\small 10}};
\draw (124.88,165.58) node [anchor=north west][inner sep=0.75pt]   [align=left] {{\tiny SL:AMOUNT}};
\draw (151.75,63.96) node [anchor=north west][inner sep=0.75pt]  [font=\footnotesize] [align=left] {{\small percent}};
\draw (172.35,143.93) node [anchor=north west][inner sep=0.75pt]  [font=\footnotesize] [align=left] {{\small por}};
\draw (201.76,165.32) node [anchor=north west][inner sep=0.75pt]   [align=left] {{\tiny SL:UNIT}};
\draw (158.01,43.32) node [anchor=north west][inner sep=0.75pt]   [align=left] {{\tiny SL:UNIT}};
\draw (102.11,43.82) node [anchor=north west][inner sep=0.75pt]   [align=left] {{\tiny SL:AMOUNT}};
\draw (200.85,142.93) node [anchor=north west][inner sep=0.75pt]  [font=\footnotesize] [align=left] {ciento};
\draw (26.5,7.5) node [anchor=north west][inner sep=0.75pt]   [align=left] {Attention Alignment};
\draw (298.9,66.17) node [anchor=north west][inner sep=0.75pt]  [font=\footnotesize] [align=left] {{\small set}};
\draw (328.37,65.17) node [anchor=north west][inner sep=0.75pt]  [font=\footnotesize] [align=left] {{\small volume}};
\draw (378.34,64.17) node [anchor=north west][inner sep=0.75pt]  [font=\footnotesize] [align=left] {{\small to}};
\draw (403.91,65.17) node [anchor=north west][inner sep=0.75pt]  [font=\footnotesize] [align=left] {{\small 10}};
\draw (289.4,142.93) node [anchor=north west][inner sep=0.75pt]  [font=\footnotesize] [align=left] {{\small ajuste}};
\draw (330.82,142.78) node [anchor=north west][inner sep=0.75pt]  [font=\footnotesize] [align=left] {{\small el}};
\draw (352.17,142.93) node [anchor=north west][inner sep=0.75pt]  [font=\footnotesize] [align=left] {{\small volumen}};
\draw (404.67,143.93) node [anchor=north west][inner sep=0.75pt]  [font=\footnotesize] [align=left] {{\small a}};
\draw (425.66,143) node [anchor=north west][inner sep=0.75pt]  [font=\footnotesize] [align=left] {{\small 10}};
\draw (403.88,165.58) node [anchor=north west][inner sep=0.75pt]   [align=left] {{\tiny SL:AMOUNT}};
\draw (430.75,63.96) node [anchor=north west][inner sep=0.75pt]  [font=\footnotesize] [align=left] {{\small percent}};
\draw (451.35,143.93) node [anchor=north west][inner sep=0.75pt]  [font=\footnotesize] [align=left] {{\small por}};
\draw (467.26,165.32) node [anchor=north west][inner sep=0.75pt]   [align=left] {{\tiny SL:UNIT}};
\draw (437.01,43.32) node [anchor=north west][inner sep=0.75pt]   [align=left] {{\tiny SL:UNIT}};
\draw (381.11,43.82) node [anchor=north west][inner sep=0.75pt]   [align=left] {{\tiny SL:AMOUNT}};
\draw (479.85,142.93) node [anchor=north west][inner sep=0.75pt]  [font=\footnotesize] [align=left] {ciento};
\draw (305.5,7.5) node [anchor=north west][inner sep=0.75pt]   [align=left] {Fast-align};

\end{tikzpicture}

\caption{An example comparison between the two methods of slot label projection. The image in on the left shows Attention alignment, where every source token gets projected to a single target token. As a result, \textit{percent}, in EN is aligned only with \textit{ciento} in ES. The image on the right shows fast-align, which allow a many-to-many alignment. Hence \textit{percent} is correctly aligned with \textit{por ciento}.}
\label{fig:translate}

\end{figure*}
}